\documentclass[final,1p,times,authoryear]{elsarticle}

\usepackage{amsmath}
\usepackage{amsfonts}
\usepackage{natbib}
\usepackage{rotating}
\usepackage{float}
\usepackage[linesnumbered,ruled,vlined]{algorithm2e}

\usepackage[utf8]{inputenc} 
\usepackage[T1]{fontenc}    

\usepackage{url}            
\usepackage{booktabs}       
\usepackage{amsfonts}       
\usepackage{nicefrac}       
\usepackage{microtype}      
\usepackage{xcolor}         

\usepackage{hyperref} 
\hypersetup{
    colorlinks = true,
    linkcolor = red,
    anchorcolor = blue,
    citecolor = blue,
    filecolor = blue,
    urlcolor = blue
    }

\journal{Computers in Biology and Medicine}

\begin{document}

\begin{frontmatter}

\title{From Motion to Meaning:\\Biomechanics-Informed Neural Network for Explainable Cardiovascular Disease Identification}


\author[bcn,jrcgeel]{Valentin Comte}
\author[bcn,jrcispra]{Mario Ceresa}
\author[bcn]{Gemma Piella}
\author[bcn,icrea]{Miguel A. González Ballester}

\affiliation[bcn]{organization={BCN MedTech, Department of Engineering, Universitat Pompeu Fabra},
            addressline={Carrer de Roc Boronat, 138, Sant Martí}, 
            city={Barcelona},
            postcode={08018},
            country={Spain}}

\affiliation[jrcgeel]{organization={European Commission, Joint Research Centre (JRC)},
            addressline={Retieseweg 111}, 
            city={Geel},
            postcode={2440}, 
            country={Belgium}}

\affiliation[jrcispra]{organization={European Commission, Joint Research Centre (JRC)},
            addressline={JRC Ispra, 1}, 
            city={Ispra},
            postcode={21027},
            country={Italy}}

\affiliation[icrea]{organization={ICREA},
            addressline={Passeig de Lluís Companys, 23}, 
            city={Barcelona},
            postcode={08010},
            country={Spain}}

\begin{abstract}
Cardiac diseases are among the leading causes of morbidity and mortality worldwide, which requires accurate and timely diagnostic strategies. In this study, we introduce an innovative approach that combines deep learning image registration with physics-informed regularization to predict the biomechanical properties of moving cardiac tissues and extract features for disease classification. We utilize the energy strain formulation of Neo-Hookean material to model cardiac tissue deformations, optimizing the deformation field while ensuring its physical and biomechanical coherence. This explainable approach not only improves image registration accuracy, but also provides insights into the underlying biomechanical processes of the cardiac tissues. Evaluation on the Automated Cardiac Diagnosis Challenge (ACDC) dataset achieved Dice scores of 0.945 for the left ventricular cavity, 0.908 for the right ventricular cavity, and 0.905 for the myocardium. Subsequently, we estimate the local strains within the moving heart and extract a detailed set of features used for cardiovascular disease classification.  We evaluated five classification algorithms — Logistic Regression, Multi-Layer Perceptron, Support Vector Classifier, Random Forest, and Nearest Neighbour — and identified the most relevant features using a feature selection algorithm. The best performing classifier obtained a classification accuracy of 98\% in the training set and 100\% in the test set of the ACDC dataset. By integrating explainable artificial intelligence, this method empowers clinicians with a transparent understanding of the model's predictions based on cardiac mechanics, while also significantly improving the accuracy and reliability of cardiac disease diagnosis, paving the way for more personalized and effective patient care.
\end{abstract}

\begin{keyword}

Medical image registration \sep Deep learning \sep Biomechanics \sep Cardiovascular diseases \sep Healthcare \sep Classification



\end{keyword}

\end{frontmatter}

\section{Introduction}

Cardiovascular diseases (CVDs) remain the leading cause of death worldwide, accounting for nearly 17.9 million deaths each year, which represents approximately 31\% of all global deaths \citep{nawaz2021intelligent}. The early and accurate diagnosis of CVDs is crucial for effective treatment and management, potentially reducing the mortality rate and improving patient outcomes. However, the traditional methods of diagnosing CVDs, which often rely on visual interpretation of medical images and clinical data, are time-consuming, subject to inter-observer variability, and can be prone to errors. In this context, the advent of deep learning (DL) technologies has opened new avenues for the automatic classification of CVDs. DL algorithms, particularly convolutional neural networks (CNNs), have demonstrated remarkable performance in medical image analysis, often surpassing human experts in accuracy.\\

\noindent
Cardiac disease classification is a rapidly evolving field, with advances in machine learning and DL techniques significantly improving the accuracy and efficiency of disease detection and classification. The integration of DL for the classification of CVDs offers several significant advantages. Firstly, it enables the rapid and consistent analysis of vast amounts of data, facilitating early detection and intervention. Secondly, it reduces the workload on healthcare professionals, allowing them to focus on patient care rather than repetitive diagnostic tasks. Additionally, automated classification systems can be continuously updated and improved as new data becomes available. This ensures that diagnostic accuracy is maintained over time and generalizes well across images acquired with different scanners, which may exhibit heterogeneous appearances. In that context, cine-cardiac magnetic resonance imaging (MRI) constitutes a valuable asset because of its ability to capture subtle motions of the cardiac tissues. Several cine-cardiac MRI datasets are available for research purposes, each offering different advantages and limitations. The ACDC dataset \citep{bernard2018deep} is one of the most widely used benchmarks. It consists of 2D short-axis cine-MRI slices stacked to reconstruct the left ventricle volume, with expert annotations and pathology labels for 150 patients. It supports both segmentation and classification of five cardiac conditions. The UK Biobank (UKBB) dataset \citep{petersen2016reference} follows a similar 2D acquisition scheme but includes a much larger and more diverse population, making it ideal for large-scale studies. In contrast, the dataset used in \cite{galazis2023high} consists of true 3D cine-MRI volumes of the left atrium acquired over time, allowing for anatomically detailed biomechanical strain analysis. However, this dataset is not publicly available and includes only a small number of subjects collected at a single centre. Building on the ACDC dataset, several studies have explored automatic CVD classification using machine learning. \cite{isensee2018automatic} extracted features from segmentations and trained ensembles of MLPs and random forests, achieving classification accuracies of 94\% (training) and 92\% (test). \cite{cetin2018radiomics} used radiomics and SVMs with feature selection, reaching 96\% and 92\% accuracy, while \cite{Khened2017DenselyCF} proposed a DenseNet-based segmentation pipeline coupled with random forest classification, achieving 90\% and 96\%. Although effective, these approaches rely primarily on static anatomical features and do not leverage dynamic or biomechanical information from cine-MRI sequences.

\noindent
This gap can be effectively bridged using deformable image registration (DIR), which computes a nonlinear deformation field (DF) to align pairs of images. DIR is particularly advantageous for the dynamic analysis of moving tissues, offering precise approximations of particle DFs. By incorporating biomechanical priors to constrain the DF, more realistic and physiologically plausible deformations can be achieved. This approach is especially useful in analyzing the dynamic motion of the heart using cine-cardiac MRI sequences, enabling the detection of biomechanical anomalies. Such analysis plays a crucial role in diagnosing pathologies that disrupt normal cardiac tissue motion, with the added benefit of enhancing the explainability of artificial intelligence (AI) models in clinical settings. Biomechanical analysis of cardiac tissues involves examining the mechanical properties and behavior of the heart under various physiological conditions. This includes understanding the stiffness, elasticity, and motion of cardiac tissues, providing valuable insights into heart health and functionality. Since biomechanical properties can be significantly altered in various cardiac pathologies, this analysis is essential for diagnosing and treating such conditions. Advanced techniques, such as biomechanics-informed neural networks and physics-informed neural networks, have been developed to track myocardial motion and register cine-MRI scans accurately. By integrating biomechanical insights, these approaches make the diagnostic process more transparent and reliable, allowing clinicians to understand and trust the AI decision-making process. This transparency is crucial for the adoption of AI in medical practice, ensuring that the models provide clear, interpretable results that can be effectively used in patient care.\\

\noindent
\cite{qin2020biomechanics} introduced an innovative method to monitor myocardial movements by embedding a biomechanics-informed prior into their registration model, rather than using an explicit generic regularization. This approach learns a manifold for biomechanically plausible deformations. Their method was evaluated on the UK biobank (UKBB) \citep{petersen2016reference} and ACDC datasets. By registering the 2D slices of the ES frames onto the ED frames, they achieved Dice scores of 0.684 for the apical slices, 0.796 for the mid-ventricle slices, and 0.725 for the basal slices. While their method outperformed traditional registration methods with conventional regularization, it was not compared to SOTA DL registration methods. In a similar inspiration, \cite{wu2024neural} utilised a registration framework that integrates neural differential equations to model DFs, learning dynamics directly from data without relying on physical priors. By registering the ES onto the ED frames of the ACDC dataset, they reported Dice scores of 0.702, 0.788, and 0.793 for the apical, mid-ventricle, and basal slices, respectively. Although their approach achieved promising results, it performed slightly lower than the SOTA DL registration methods they reported. Biomechanics-inspired regularizations can also be used to constrain the displacement vector fields (DVF), as proposed by \cite{zhang2022learning}, with a biomechanics-informed prior as regularisation on predicted DVFs, thereby enhancing the plausibility of the transformation. They tested their approach on 2D slices of the ACDC dataset and reported Dice scores of 0.927, 0.859, and 0.925 for the left ventricle (LV), right ventricle (RV), and epicardium, respectively. \cite{galazis2023high} introduced a tool for the  analysis of left atrial displacements and deformations using unsupervised neural networks (Aladdin), designed to automatically characterise regional left atrial deformations from 3D cine MRI scans, which includes online segmentation and unsupervised image registration networks, as well as a strain calculation pipeline. The study applied Aladdin to data from healthy volunteers and patients with CVDs, generating maps of the magnitude of the left atrial DVF and the principal strain values. While these methods show promising results by using biomechanical priors to constrain DVFs and estimating strains, no study to date has utilized these strains for the automatic classification of CVDs. Table \ref{tab:cine_summary_red} provides a comprehensive summary of recent studies focused on cine cardiac MRI analysis.
\\

\begin{table}[H]
\centering
\scriptsize
\begin{tabular}{|p{2cm}|p{1.5cm}|p{1.5cm}|p{2.5cm}|p{2.5cm}|p{1cm}|}
\hline
\textbf{Paper} & \textbf{Modality} & \textbf{Application} & \textbf{Details} & \textbf{Limitations} & \textbf{Dataset} \\
\hline
\cite{isensee2018automatic} & stacked 2D Cine MRI & Segmentation, classification & Ensemble of 2D and 3D U-Nets on ACDC dataset, classification based on extracted features using MLP and random forest & Limited generalization to other datasets, performance dependent on manual annotations & ACDC \\
\hline
\cite{Khened2017DenselyCF} & stacked 2D Cine MRI & Segmentation, classification & Ensemble CNN (U-Net) for segmentation, feature-based classification & Limited generalization to other datasets, performance dependent on manual annotations & ACDC \\
\hline
\cite{cetin2018radiomics} & stacked 2D Cine MRI & Classification & Radiomics features + SVM classification on ACDC dataset & Limited generalization to other datasets, performance dependent on manual annotations, radiomics sensitive to acquisition variability & ACDC \\
\hline
\cite{qin2020biomechanics} & 2D Cine MRI & Registration, Biomechanical Analysis & Biomechanical constraints using VAE latent space, better generalization & Limited realism without full 3D, no classification & UKBB \\
\hline
\cite{zhang2022learning} & 2D Cine MRI & Motion Tracking, Biomechanical Analysis & Biomechanics-informed correspondences learned via graph modeling & Limited spatial resolution, early-stage pipeline, no classification & ACDC \\
\hline
\cite{wu2024neural} & stacked 2D Cine MRI & Registration, Segmentation Propagation & Sequential registration method for better label propagation & Limited to propagation use case, no classification & ACDC \\
\hline
\cite{galazis2023high} & 3D Cine MRI & Segmentation, Registration, Biomechanical Analysis & Left atrial segmentation (nnU-Net), registration (Aladdin-R), strain atlas creation & Small dataset, strain computed on LA only, no classification & Private \\
\hline

\end{tabular}
\caption{Summary of cine cardiac MRI analysis studies,}
\label{tab:cine_summary_red}
\end{table}

\noindent

This study addresses two major limitations in the current state of the art: existing classification methods rely predominantly on static or radiomic features and fail to exploit the biomechanical dynamics of the heart, while recent biomechanics-informed registration approaches, despite estimating meaningful deformation fields and strain maps, have not been used for automated disease classification. To bridge this gap, we propose a deep learning framework that combines deformable image registration with physics-informed regularisation based on Neo-Hookean strain energy, enabling the extraction of physiologically plausible deformation fields from cine-MRI sequences. From these, we derive local strain and mechanical properties such as shear and bulk moduli, which are integrated with radiomic and volumetric features to classify the five cardiac pathologies in the ACDC dataset. By capturing both anatomical structure and dynamic tissue behaviour, the proposed method enhances diagnostic accuracy while providing interpretable, physiology-based markers that support clinical trust and adoption.

\section{Methods}\label{method}

\subsection{Cascaded registration}

\begin{figure}[h!]
\centering
\includegraphics[width=0.95\textwidth]{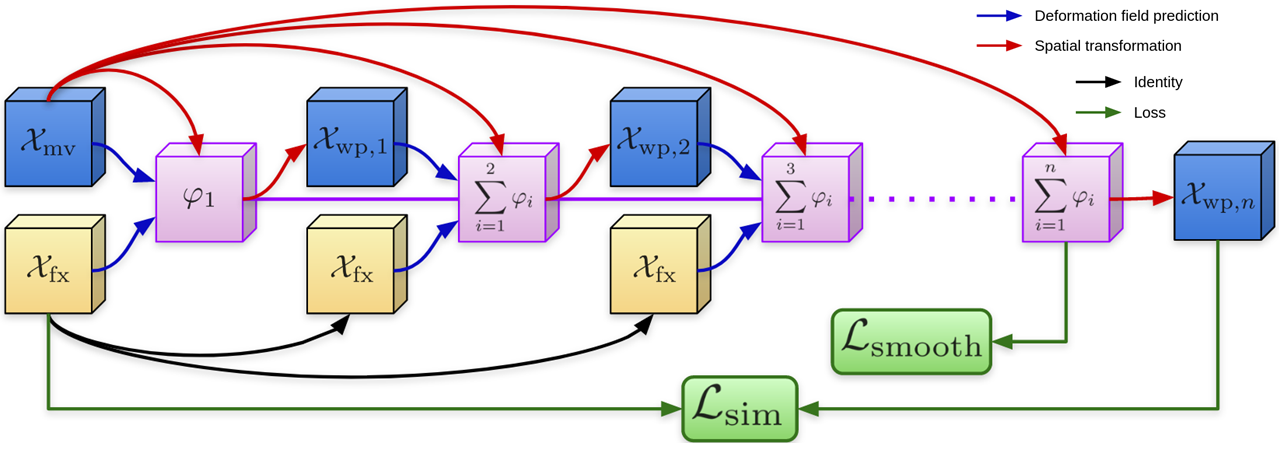}\hfill
\caption{Overview of the cascaded registration model, adapted from \cite{comte2025deep}.}
\label{fig:casreg}
\end{figure}

In our earlier work \citep{comte2025deep}, we developed a cascaded DL model (Figure \ref{fig:casreg}) designed for the registration of 3D medical images. This model breaks down the registration process into sequential steps, using multiple DFs to achieve alignment at different spatial scales.

\noindent
Registering two 3D images involves aligning a moving image \(\mathcal{X}_\mathrm{mv}\) with a fixed image \(\mathcal{X}_\mathrm{fx}\) over a 3-dimensional domain \(\Omega \subset \mathbb{R}^3\). The goal is to find a spatial transformation \(\varphi: \mathbb{R}^3 \rightarrow \mathbb{R}^3\) that warps \(\mathcal{X}_\mathrm{mv}\) to resemble \(\mathcal{X}_\mathrm{fx}\):

\begin{equation}
    \mathcal{X}_\mathrm{wp} =  \mathcal{X}_\mathrm{mv} \circ \varphi \approx \mathcal{X}_\mathrm{fx} .
    \label{registration_eq}
\end{equation}

\noindent
Traditional registration techniques typically involve multiple sequential stages, beginning with affine registration and advancing through deformable image registration (DIR) using a coarse-to-fine approach. This multi-stage process has been replicated in DL-based registration, where stacked CNNs progressively refine image alignment \citep{de2019deep}. In contrast to these sequential multi-stage models, we proposed a cascaded model in which stacked networks process the inputs simultaneously, with the loss function computed only on the final output, ensuring a collaborative behaviour of the cascaded networks. Our model is composed of a sequence of CNNs, each predicting an incremental deformation field \(\varphi_i\) that progressively aligns the moving image to the fixed image. The outputs of the networks are not applied sequentially to warp the image repeatedly—instead, the predicted deformation fields are accumulated and applied once at the end of the cascade, thereby avoiding repeated interpolation and reducing information loss. Each CNN operates at a different resolution scale, allowing the model to capture both coarse global transformations and fine local deformations. This multi-resolution design is reinforced by a multi-scale similarity loss that calculates normalized cross-correlation (NCC) over patches of different sizes, promoting alignment across spatial frequencies. The registration process can be summarized as follows:

\begin{enumerate}
  \item The first network inputs \(\mathcal{X}_\mathrm{mv}\) and \(\mathcal{X}_\mathrm{fx}\) to produce a dense DF, \(\varphi_1\), partially aligning the moving image with the fixed image.
  \item The resulting warped image, \(\mathcal{X}_\mathrm{wp,1}\), is input to the second network with \(\mathcal{X}_\mathrm{fx}\) to generate \(\varphi_2\). This DF is summed with \(\varphi_1\) to further warp \(\mathcal{X}_\mathrm{mv}\) into \(\mathcal{X}_\mathrm{wp,2}\).
  \item This process repeats recursively, resulting in a final DF that fully aligns the moving image with the fixed image:
  \begin{align}
    \begin{split}
        \mathcal{X}_\mathrm{wp,n} = \mathcal{X}_\mathrm{mv} \circ \sum_{i=1}^n  \varphi_i .
    \end{split}
  \end{align}
\end{enumerate}

\noindent
We adopt the loss function proposed by \cite{Voxelmorph}, which blends an image similarity loss with a regularization term. The similarity loss, $\mathcal{L}_\mathrm{sim}$, is formulated as the negative local cross-correlation, designed to enhance the alignment between the warped and fixed images:

\begin{align}
    \mathcal{L}_\mathrm{sim} = -CC(\mathcal{X}_\mathrm{fx},\mathcal{X}_\mathrm{wp}) = - \sum_{p \in \Omega} \frac{\left(\sum_{p_i} (\mathcal{X}_\mathrm{fx}(p_i) - \bar{\mathcal{X}}_\mathrm{fx}(p)) (\mathcal{X}_\mathrm{wp}(p_i)-\bar{\mathcal{X}}_\mathrm{wp}(p))\right)^2}{\left(\sum_{p_i}(\mathcal{X}_\mathrm{fx}(p_i) - \bar{\mathcal{X}}_\mathrm{fx}(p))^2\right)\left(\sum_{p_i}(\mathcal{X}_\mathrm{wp}(p_i)-\bar{\mathcal{X}}_\mathrm{wp}(p))^2\right)} ,
\end{align}

\vspace{.5cm}
\noindent
where \(p_i\) iterates over the volume \(d^3\) around voxel \(p\), and \(\bar{\mathcal{X}}_\mathrm{fx}(p)\) and \(\bar{\mathcal{X}}_\mathrm{wp}(p)\) are local mean intensities over \(d^3\) for the fixed and warped images. This loss is computed for patches of varying sizes, promoting image similarity at multiple scales.

\subsection{Neo-Hookean model implementation}

\noindent
To regularize the DF $\varphi$, a common approach is to penalize the sum of its local gradients, $\sum_{p \in \Omega} || \nabla \varphi(p)||^2$, ensuring a smoother and more invertible transformation. However, this purely mathematical regularization does not account for the biomechanical properties of cardiac tissues. This lack of biomechanical consideration makes the model less interpretable in a medical context. Therefore, we introduce a novel regularization term based on the Neo-Hookean energy strain, which enforces biophysically plausible deformations during the registration process. This approach not only maintains the smoothness and invertibility of the transformation but also enhances the explainability of the model by aligning with known biomechanical behaviors of human tissues. Human tissues exhibit complex nonlinear stress-strain behavior, and cardiac tissue, in particular, demonstrates such nonlinear characteristics. In this study, we approximate the heart muscle as a hyperelastic material and use the Neo-Hookean model to describe its energy strain for simplicity.\\

\subsubsection{Neo-Hookean Energy Strain}

In the context of 3D Euclidean space, denoted by $\mathbb{R}^3$, a solid body is described as a collection of interacting particles. The motion of the solid body is defined as the evolution of its configuration, resulting in a new position for each particle at time $t$, denoted by:

\begin{equation}
    \mathrm{\bf{x}}^\varphi = \varphi (\mathrm{\bf{x}}) .
\end{equation}

\noindent
The position vectors for each particle in its initial and deformed configurations are represented as:

\begin{equation}
    \mathrm{\bf{x}} = x_i e_i,~\mathrm{and}~~ \mathrm{\bf{x}}^\varphi = x_i^\varphi e_i ,
\end{equation}

\noindent
where $e_i$ are the basis vectors of $\mathbb{R}^3$. We can then define the function $u(\mathrm{\bf{x}}) = u_i e_i$, describing the relative displacement of a particle:

\begin{equation}
    u(\bf{x}) = \bf{x}^\varphi - \bf{x} = \varphi(\bf{x}) - \bf{x} .
\end{equation}

\noindent
To describe the deformation of the solid body, we introduce the deformation gradient $\mathbf{F}$, which is defined as the gradient of the DF $\varphi$ with respect to the initial configuration $\mathbf{x}$:

\begin{align}
    \mathbf{F}(\varphi) = \dfrac{\partial \varphi(\bf{x})}{\partial \mathbf{x}}& = \begin{pmatrix}
    \dfrac{\partial x_1^\varphi}{\partial x_1} & \dfrac{\partial x_1^\varphi}{\partial x_2} & \dfrac{\partial x_1^\varphi}{\partial x_3} \\ \\
    \dfrac{\partial x_2^\varphi}{\partial x_1} & \dfrac{\partial x_2^\varphi}{\partial x_2} & \dfrac{\partial x_2^\varphi}{\partial x_3} \\ \\
    \dfrac{\partial x_3^\varphi}{\partial x_1} & \dfrac{\partial x_3^\varphi}{\partial x_2} & \dfrac{\partial x_3^\varphi}{\partial x_3}
    \end{pmatrix} =
    \begin{pmatrix}
    \dfrac{\partial u_1}{\partial x_1} + 1 & \dfrac{\partial u_1}{\partial x_2} & \dfrac{\partial u_1}{\partial x_3} \\ \\
    \dfrac{\partial u_2}{\partial x_1} & \dfrac{\partial u_2}{\partial x_2} + 1 & \dfrac{\partial u_2}{\partial x_3} \\ \\
    \dfrac{\partial u_3}{\partial x_1} & \dfrac{\partial u_3}{\partial x_2} & \dfrac{\partial u_3}{\partial x_3} + 1 \\
    \end{pmatrix} .
    \label{def_grad_tens}
\end{align}

\noindent
The deformation gradient is a two-point tensor as it relates infinitesimal displacements or line elements in the undeformed configuration with corresponding displacements in the deformed configuration. It captures local elongations as well as rotations. The deformation gradient $\mathbf{F}$ can be decomposed into the stretch tensors $\mathbf{U}$ and $\mathbf{V}$ and a rotation tensor $\mathbf{R}$ as follows:

\begin{align}
    F & = \mathbf{R} \mathbf{U} = \mathbf{V} \mathbf{R} .\\
\end{align}

\noindent
From the deformation gradient, the left and right Cauchy-Green deformation tensors can be derived:

\begin{align*}
    & \mathbf{B} = \mathbf{F}\mathbf{F}^T , \\
    & \mathbf{C} = \mathbf{F}^T\mathbf{F} ,
\end{align*}

\noindent
where $\mathbf{B}$ is the left and $\mathbf{C}$ is the right Cauchy-Green deformation tensor. Both tensors are independent of the rotation and only measure local stretches. Thus, the eigenvalues of $\mathbf{U}$ and $\mathbf{V}$ are called the principal stretches $\lambda_i$ and the eigenvalues of $\mathbf{B} = \mathbf{F}\mathbf{F}^T$ and $\mathbf{C} = \mathbf{F}^T\mathbf{F}$ are $\lambda_i^2$.

\begin{align}
\mathbf{I}_{1C} &= \mathrm{tr}(\mathbf{C}) , \\
\mathbf{I}_{2C} &= \dfrac{1}{2} \left( (\mathrm{tr}(\mathbf{C}))^2 - \mathrm{tr}(\mathbf{C}^2) \right) , \\
\mathbf{I}_{3C} &= \det(\mathbf{C}) .
\end{align}

\noindent
These invariants play a crucial role in the mechanics of deformable solids. For the description of an isotropic homogeneous incompressible material the
definition of $\mathbf{I}_{1C}$ and $\mathbf{I}_{2C}$ are sufficient. For compressible materials $\mathbf{I}_{3C}$ also has to be taken into consideration. The volume change of a volume element is defined by the determinant of the deformation gradient tensor:

\begin{equation}
    \mathbf{J} = det(\mathbf{F}) .
\end{equation}

\noindent
However, the deformation gradient $\mathbf{F}$ captures the whole deformation of a material and does not allow a direct differentiation between a distortional and a volumetric component. This differentiation however is necessary when incompressible or nearly incompressible materials are considered. Following \cite{bonet2000finite}, the volumetric change captured by the deformation gradient can be achieved by multiplying $\mathbf{F}$ with $\mathbf{J}^{-1/3}$:

\begin{equation}
    \hat{\mathbf{F}} = \mathbf{J}^{-1/3}\mathbf{F} ,
\end{equation}

\noindent
which ensure $det(\hat{\mathbf{F}}) \approx 1$. The deviatoric right Cauchy-Green deformation tensor is then defined by:

\begin{equation}
    \hat{\mathbf{C}} = \hat{\mathbf{F}}^T \hat{\mathbf{F}} = \mathbf{J}^{-2/3}  \mathbf{C} .
\end{equation}

\noindent
Hence, the first Cauchy invariant becomes:

\begin{equation}
    \hat{\mathbf{I}}_{1C} = tr(\mathbf{C}) \mathbf{J}^{-2/3} .
\end{equation}

\noindent
We can consider Neo-Hookean material, which is a type of material that exhibits a nonlinear stress-strain relationship. It is defined in terms of the first and third invariant of the right Cauchy-Green deformation tensor and is thus better equipped to also deal with large strains:

\begin{align}
\Phi(\varphi) &= \dfrac{\mu}{2} \left( \hat{\mathbf{I}}_{1C} - 3 \right) + \dfrac{\kappa}{2} \left(\mathbf{J}-1 \right)^2 ,\\
&= \dfrac{\mu}{2} \left( \mathbf{I}_{1C} J^{-2/3} - 3 \right) + \dfrac{\kappa}{2} \left(\mathbf{J}-1 \right)^2 ,\\
&= \dfrac{\mu}{2} \cdot \Phi_\mathrm{dis}(\varphi) + \dfrac{\kappa}{2} \cdot \Phi_\mathrm{vol}(\varphi),
\label{nhe_eq}
\end{align}

\noindent
where $\mu$ and $\kappa$ are material constants, the shear and bulk modulus, repectively. The first term $\frac{\mu}{2} \cdot \Phi_\mathrm{dis}(\varphi)$ captures the distortional changes of the material, while the second term $\frac{\kappa}{2} \cdot \Phi_\mathrm{vol}(\varphi)$ describes the volumetric changes.\\ 

\subsubsection{Biomechanical regularization}

The DF $\varphi$ predicted by our model is crucial for modeling the actual physical deformation of cardiac tissues during the registration process. From this DF, we derive the deformation gradient tensor, as described in Equation \ref{def_grad_tens}, which in turn is used to compute the Neo-Hookean energy strain (Equation \ref{nhe_eq}) for the deformed cardiac tissues. The final loss function, composed of the image similarity term and the biomechanical regularization is described as follows:

\begin{align}
    \mathcal{L} &= \mathcal{L}_\mathrm{sim} + \lambda \cdot \mathcal{L}_\mathrm{nhe} ,\\
    &= -CC(\mathcal{X}_\mathrm{fx},\mathcal{X}_\mathrm{wp}) + \lambda \cdot \Phi(\varphi) ,
\end{align}

\noindent
where $\lambda$ is the regularization parameter. Incorporating the Neo-Hookean energy strain into our regularization strategy serves a dual purpose. Firstly, it ensures that the predicted deformations are physically plausible, describing more accurately the biomechanical behavior of cardiac tissues. Secondly, this regularization helps the network produce smoother and more invertible DFs by penalizing unrealistic distortions or volumetric changes. Thus, the Neo-Hookean energy strain model acts as a biomechanically informed constraint, guiding the network to generate deformations that are both physically and functionally coherent. 

\subsubsection{Estimation of the biomechanical parameters}

For cardiac tissues, which can be approximated as nearly incompressible materials, the ratio of bulk modulus to shear modulus $\kappa/\mu$ is typically large. In this model, we have assumed a ratio of $\kappa/\mu = 50$. Given an estimated shear modulus value of $\mu = 2~\text{kPa}$ as reported in previous studies \citep{pislaru2014viscoelastic}, the corresponding bulk modulus can be approximated as $\kappa = 100~\text{kPa}$. This configuration reflects the mechanical behavior of cardiac tissues, which strongly resist compression but accommodate shear. Accordingly, our network is regularized using the full neo-Hookean energy function, which simultaneously penalizes volumetric and distortional deformations. This was intentionally designed to enforce the physical plausibility of the predicted deformation fields. By promoting nearly volume-preserving transformations while allowing shape adaptation, the regularization improves anatomical realism and stability during training. Using these approximations, we can further estimate the local shear and bulk moduli in each voxel $p$, denoted as $\mu(p)$ and $\kappa(p)$. Firstly, we compute the local DF for each time frame by registering with the preceding and following frames: 

\begin{align}
    \varphi(p,t) = \dfrac{1}{2} \cdot \left( \varphi(p,t-1 \rightarrow t) + \varphi(p, t \rightarrow t+1) \right) ,
\end{align}

\noindent
where $\varphi(p, t-1 \rightarrow t)$ denotes the local DF between frame $t-1$ and $t$, and $\varphi(p, t \rightarrow t+1)$ the local DF between $t$ and $t+1$. We can then approximate the local bulk and shear moduli:

\begin{align}
    \mu(p,t) \cdot \Phi_\mathrm{dis}(p, \varphi(p,t)) = \mu \cdot \sum_{p_i} \frac{\Phi_\mathrm{dis}(p_i, \varphi(p_i, t))}{d^3} ,\\
    \kappa(p,t) \cdot \Phi_\mathrm{vol}(p, \varphi(p,t)) = \kappa \cdot \sum_{p_i} \frac{\Phi_\mathrm{vol}(p_i, \varphi(p_i,t))}{d^3} ,
\end{align}

\noindent
where \(p_i\) iterates over the volume \(d^3\) around voxel \(p\). Hence we obtain:

\begin{align}
    \mu(p) = \dfrac{\mu \cdot \sum_{p_i} \Phi_\mathrm{dis}(p_i, \varphi(p_i))}{d^3 \cdot \Phi_\mathrm{dis}(p, \varphi(p))}, \quad \kappa(p) = \dfrac{\kappa \cdot \sum_{p_i} \Phi_\mathrm{vol}(p_i, \varphi(p_i))}{d^3 \cdot \Phi_\mathrm{vol}(p, \varphi(p))} .
\end{align}

\noindent
These estimates allow for the biomechanical properties of the tissue to be inferred at voxel level, providing a more accurate representation of cardiac mechanics.

\subsection{Registration evaluation}\label{registration_eval}

We evaluated the performance of our biomechanically constrained registration model by aligning the ES and ED phases. Subsequently, we propagated the corresponding ground-truth labels and calculated the Dice similarity coefficient between the deformed and ground-truth labels. In a following phase, we adopted a multi-frame approach inspired by multi-atlas segmentation, which involves registering multiple atlas images to a target image and merging the propagated labels to achieve a refined segmentation. Instead of employing atlases, we propagated the labels of the ES and ED frames onto the adjacent frames, utilising these labels as atlases. The adjacent labelled frames of the ED or ES frames were then propagated onto the ES/ED frame and merged using Local Weighting Voting, which assigns greater weight to labels that correspond to regions of high local similarity between the adjacent frames and the target frame. 

\subsection{Classification}

\begin{figure}[!h]
\centering
\includegraphics[width=\textwidth]{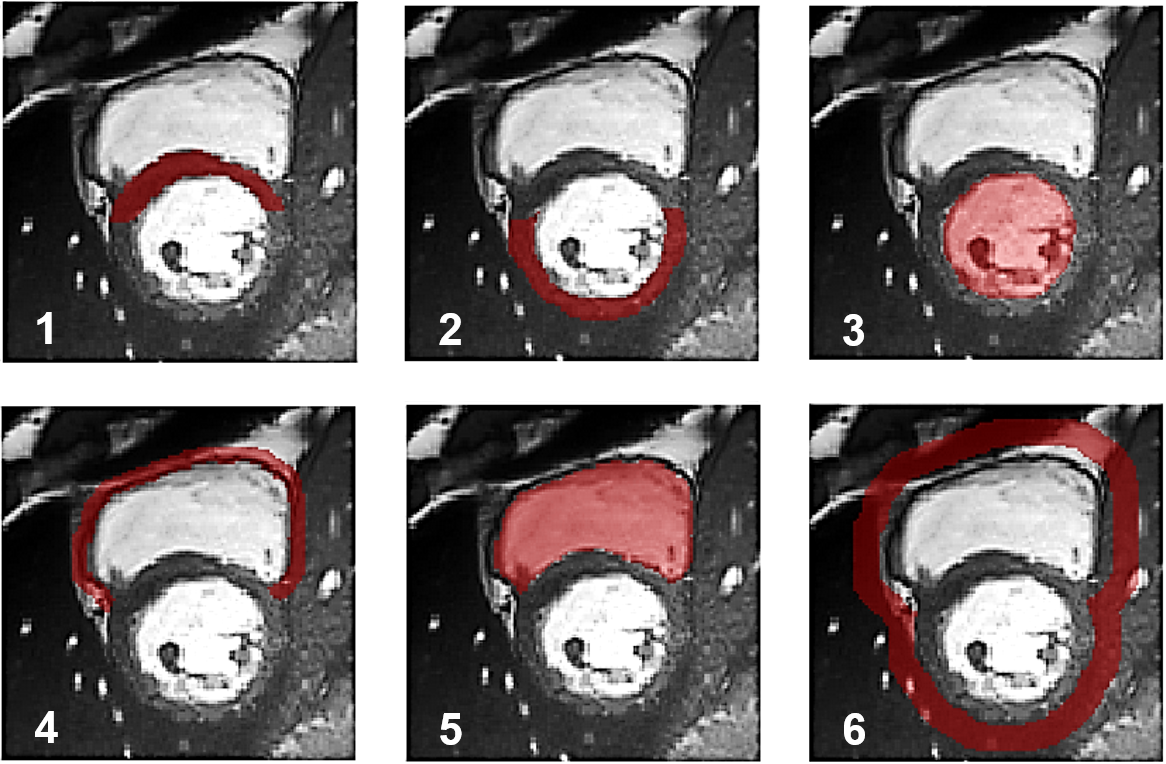}\hfill
\caption{Labels used to compute the local biomechanical features: 1) right segment of the left ventricular myocardium, 2) left segment of the left ventricular myocardium, 3) left ventricular cavity, 4) right ventricular myocardium, 5) right ventricular cavity, and surrounding pericardial tissues.}
\label{fig_labels}
\end{figure}

From the locally evaluated biomechanical constants $\mu$ and $\kappa$ at ED and ES, we extract a large set of features based on their mean values, standard deviation, 10th and 90th percentile within 6 anatomical labels, as shown in Figure \ref{fig_labels}. The set of features is further enhanced with volumetric features extracted from the ground-truth segmentation at ED and ES. 

\noindent
In order to select the most relevant features for the accurate classification of the CVDs, we implement a greedy forward-backward feature selection algorithm (Algorithm \ref{alg_feature_selection}). 

\begin{algorithm}
\caption{Feature Selection Algorithm}
\label{alg_feature_selection}
\SetAlgoLined
\KwIn{Dataset $F$ with $n$ features}
\KwOut{Selected features for classification}
Initialize discarded features set $D$ as empty\;
Initialize maximum classification accuracy $acc_{max} = 0$\;
\While{classification accuracy improves}{
    \For{$f$ in $F$}{
        $F \leftarrow F \setminus \{f\}$;
        Train classification on $F$\;
        Compute classification accuracy $acc$\;
        \If{$acc \geq acc_{max}$}{
            $F \leftarrow F \setminus \{f\}$\;
            $D \leftarrow D \cup \{f\}$\;
            $acc_{max} \leftarrow acc$\;
        }
        \Else{
            $F \leftarrow F \cup \{f\}$\;
        }
        
    }
}
\While{classification accuracy improves}{
    \For{$f$ in $D$}{
        $F \leftarrow F \cup \{f\}$\;
        Train classification model on $F$\;
        Compute classification accuracy $acc$\;
        \If{$acc \geq acc_{max}$}{
            $F \leftarrow F \cup \{f\}$\;
            $D \leftarrow D \setminus \{f\}$\;
            $acc_{max} \leftarrow acc$\;
        }
        \Else{
            $F \leftarrow F \setminus \{f\}$\;
        }
    }
}
\end{algorithm}

\section{Results}

In this section, we present the results derived from the ACDC challenge dataset, which comprises a training set of 100 instances and a testing set of 50 instances. The dataset is categorized into five distinct classes: normal cases (NOR), heart failure with infarction (MINF), dilated cardiomyopathy (DCM), hypertrophic cardiomyopathy (HCM), and abnormal right ventricle (RV). The classes are equally represented, ensuring a balanced dataset for both training and testing phases.

\subsection{Registration}

Using the registration methodology outlined in Section \ref{registration_eval}, we thoroughly evaluate the performance of our proposed approach. This evaluation involves calculating the average Dice scores for two registration scenarios: aligning ED to ES frames, and vice versa. Additionally, we implement a multi-frame strategy that leverages temporal consistency across all frames. The results, presented in Table \ref{reg_acc}, are compared against segmentation outcomes produced by the state-of-the-art nnU-Net framework \citep{isensee2021nnu}. The multi-frame strategy outperforms simple label propagation for every anatomical label and for both ED and ES phases, demonstrating the benefit of incorporating motion continuity across time. When compared to nnU-Net, our method achieves higher Dice scores for the left ventricle (LV) and myocardium (MYO) at both ED and ES, reflecting its effectiveness in capturing the motion and deformation of the myocardium and endocardial borders. However, nnU-Net outperforms our approach for the right ventricle (RV) in both phases. This may be attributed to the more complex and variable geometry of the RV, particularly near the outflow tract and basal regions, which can be harder to capture accurately through deformation-based label propagation.\\

\begin{table}[H]
\centering
\begin{tabular}{c|ccc|ccc}
    ~ & \multicolumn{3}{c|}{ES} & \multicolumn{3}{c}{ED}\\
    \hline
    \hline
    Method & LV & RV & MYO & LV & RV & MYO\\
    \hline
    nnU-Net \citep{isensee2021nnu} &  0.968 & \textbf{0.946*} & 0.902* & 0.931* & \textbf{0.899*} & 0.919\\
    propagation (ours) &  0.970 & 0.878* & 0.920* & 0.913* & 0.804* & 0.870*\\
    multi-frame (ours) &  \textbf{0.974*} & 0.912* & \textbf{0.960*} & \textbf{0.956*} & 0.857* & \textbf{0.922}\\
    \hline
\end{tabular}
\caption{Average Dice scores on the three anatomical labels at ES and ED using registration from ED to ES frames, and vice versa, multi-frame registration, and nnU-Net. \textbf{Bold} values indicate the highest average Dice score for each label and phase. Values followed by an asterisk (*) denote a statistically significant improvement over the other methods (paired t-test, $p < 0.05$).}
\label{reg_acc}
\end{table}

\subsection{Classification}

To assess the effectiveness of incorporating biomechanical features into disease classification, we utilized a feature pool that includes values derived from local biomechanical constants $\mu$ and $\kappa$, the local modulus of the deformation field $\varphi$, and volumetric information. We then evaluated the performance of five classification algorithms—logistic regression (LR), multilayer perceptron (MLP), support vector classifier (SVC), random forest (RF), and nearest neighbor (NN)—each combined with a feature selection algorithm. The classification accuracy for each method is detailed in Table \ref{tab_res1}, with LR achieving the highest accuracy, reaching 0.98 on the training set and a perfect 1.0 on the test set. Table \ref{tab_selected_features} lists the selected features along with their relative importance, showing the classification accuracy achieved when each feature is used alone, as well as the impact on accuracy when each feature is removed from the feature pool. While volumetric features generally contribute significantly to classification accuracy, the inclusion of biomechanical constants such as $\mu$ and $\kappa$ enhances the performance of the model by providing complementary information. This underscores the critical role of biomechanical features in achieving a more precise and reliable classification of cardiac diseases, highlighting their importance in developing robust diagnostic tools.\\

\begin{table}[H]
\centering
\begin{tabular}{c|c|c}
    Classifier & Accuracy train & Accuracy test\\
    \hline
    \hline
    LR & 0.98 & 1\\
    MLP & 0.91 & 0.96\\
    SVC & 0.90 & 0.92\\
    RF & 0.89 & 0.92\\
    NN & 0.87 & 0.88\\
    \hline
\end{tabular}
\caption{Classification accuracy over the train and test set for different classification methods.}
\label{tab_res1}
\end{table}

\begin{table*}[!h]
\centering
\resizebox{\textwidth}{!}{
\begin{tabular}{c|ccc|cc|cc| |c|ccc|cc|cc}
    \multicolumn{1}{c}{~} & \multicolumn{3}{c}{~} & \multicolumn{2}{c}{\scriptsize{Validation}} & \multicolumn{2}{c}{\scriptsize{Testing}} & \multicolumn{1}{c}{~} & \multicolumn{3}{c}{~} & \multicolumn{2}{c}{\scriptsize{Validation}} & \multicolumn{2}{c}{\scriptsize{Testing}}\\
    value & label & feature & phase & w/o & alone & w/o & alone & value & label & feature & phase & w/o & alone & w/o & alone\\
    \hline
    $\mu$ & 1 & $\eta_{.10}$ & ES & 0.97 & 0.20 & 0.96 & 0.32 & $\varphi$ & 1 & $\sigma$ & ES & 0.95 & 0.30 & 0.98 & 0.36 \\
    $\mu$ & 1 & $\sigma$ & ES & 0.97 & 0.22 & 1 & 0.30 & $\varphi$ & 1 & $\eta_{.90}$ & ES & 0.97 & 0.33 & 0.98 & 0.42\\
    $\mu$ & 2 & mean & ES & 0.97 & 0.35 & 0.98 & 0.40 & $\varphi$ & 2 & $\eta_{.10}$ & ED & 0.97 & 0.35 & 0.96 & 0.34\\
    $\mu$ & 3 & $\eta_{.10}$ & ES & 0.96 & 0.29 & 0.96 & 0.38 & $\varphi$ & 1,3 & ratio & ED & 0.95 & 0.32 & 0.92 & 0.32\\
    $\mu$ & 6 & $\sigma$ & ED & 0.97 & 0.24 & 0.96 & 0.14 & $\varphi$ & 2 & ratio & ED,ES & 0.97 & 0.40 & 0.90 & 0.42\\
    $\mu$ & 6 & $\eta_{.10}$ & ES & 0.97 & 0.32 & 0.98 & 0.30 & $\varphi$ & 2,3 & ratio & ES & 0.91 & 0.36 & 0.90 & 0.30\\
    $\kappa$ & 1 & $\eta_{.90}$ & ED & 0.97 & 0.16 & 0.98 & 0.18 & $V$ & all & ratio & ES & 0.94 & 0.51 & 0.96 & 0.46\\
    $\kappa$ & 1 & $\eta_{.10}$ & ED & 0.97 & 0.18 & 0.94 & 0.18 & $V$ & 3 & - & ES & 0.94 & 0.52 & 0.96 & 0.48\\
    $\kappa$ & 4 & mean & ES & 0.97 & 0.25 & 0.96 & 0.18 & $V$ & 3 & ratio & ED,ES & 0.97 & 0.63 & 0.92 & 0.62\\
    $\kappa$ & 6 & $\sigma$ & ED & 0.97 & 0.20 & 0.96 & 0.16 & $V$ & 1,3 & ratio & ED & 0.97 & 0.60 & 0.94 & 0.62\\
    $\kappa$ & 6 & $\eta_{.90}$ & ED & 0.97 & 0.25 & 0.98 & 0.20 & $V$ & LV,RV & ratio & ES & 0.97 & 0.47 & 0.98 & 0.50\\
    \hline
    \end{tabular}}
\caption{Overview of the selected features and their impact on the classification accuracy on the validation set and testing set, both when the selected feature is removed from the feature set (w/o), or when only the selected feature is used for classification (alone).}
\label{tab_selected_features}
\end{table*}

\begin{figure}[!h]
\centering
\includegraphics[width=0.95\textwidth]{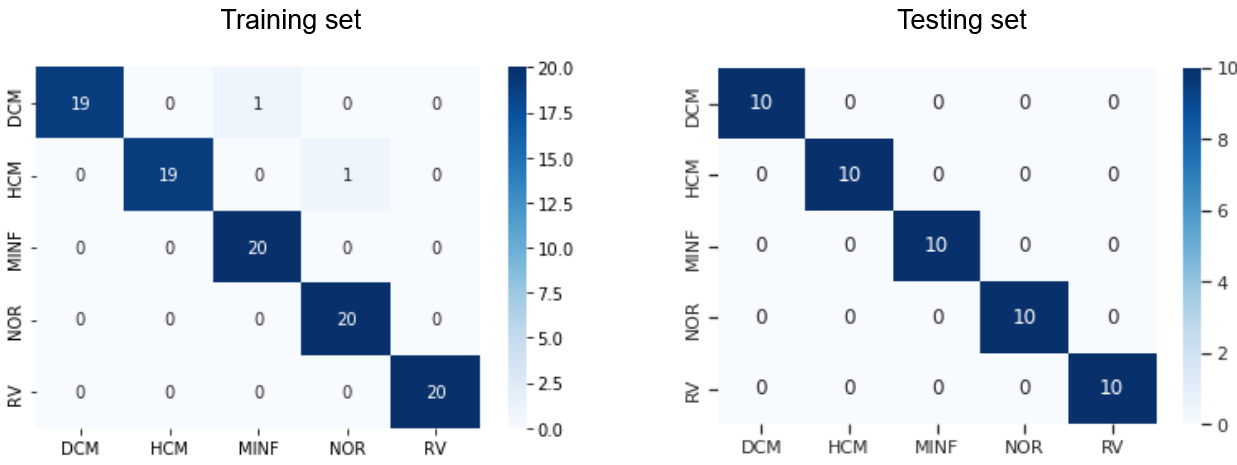}\hfill
\caption{Confusion matrices of the classification on the training set and testing set.}
\label{fig_confusion}
\end{figure}

\noindent
Figure \ref{fig_confusion} presents the confusion matrices for the classification task across five distinct cardiac disease classes. These matrices offer a detailed view of the model’s performance, showing how accurately each case is classified into its respective category. In the training set, the model performs well, with only two misclassifications out of the entire dataset. Specifically, one case of DCM is incorrectly classified as MINF, and one instance of HCM is misclassified as NOR. These misclassifications highlight the subtle overlaps in feature characteristics between these conditions, which can sometimes lead to ambiguity in the classification process. However, it is noteworthy that the model achieves perfect classification on the test set, correctly identifying all cases. This indicates that the model not only learns well from the training data but also generalizes effectively to unseen data, which is crucial for its application in real-world clinical settings.

\begin{figure}[H]
\centering
\includegraphics[width=0.9\textwidth]{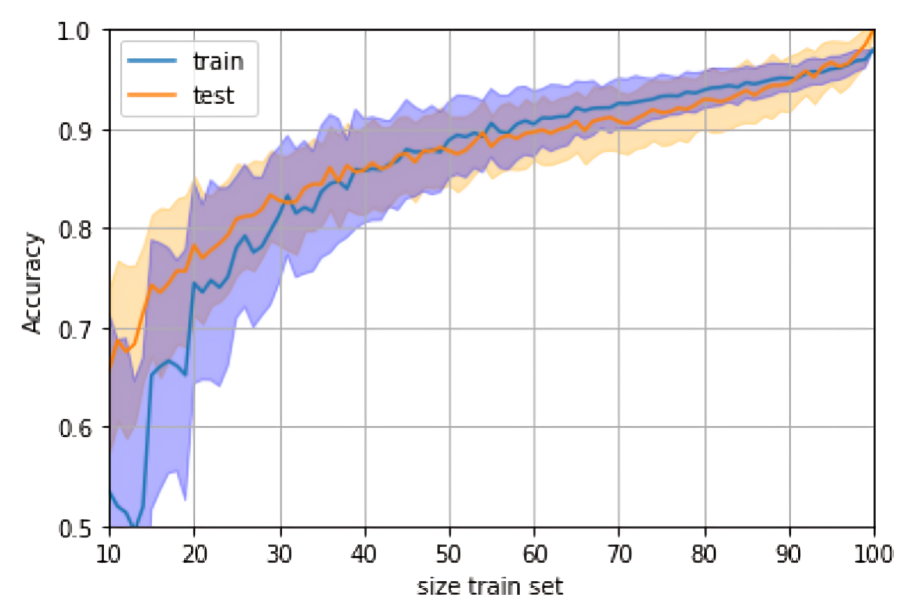}\hfill
\caption{Accuracy as a function of the training set size.}
\label{fig_generalization}
\end{figure}

Figure \ref{fig_generalization} further explores the generalization performances by illustrating the classification accuracy achieved with varying sizes of the training set. This analysis provides insight into the ability of the model to generalize as the amount of available training data decreases. The figure shows the classification accuracy obtained using the selected feature set in conjunction with the Logistic Regression (LR) classifier, plotted against different training set sizes. For each training size, fifty random subsets of cases were selected, and the average classification accuracy was computed. The solid line in the figure represents this average accuracy, while the shaded region indicates the range of one standard deviation. This analysis demonstrates the robustness of the classification model, as it maintains high classification accuracy even as the training set size is reduced, with a classification accuracy above $0.7$, $0.8$, and $0.9$ for training sets of 20, 30, and 60 subjects, respectively. The narrow shaded area, particularly at larger training set sizes, indicates low variability, suggesting that the model is stable and reliable. The maintenance of high accuracy across varying training set sizes is indicative of the model's potential to perform well even in scenarios where data availability is limited. This robustness is crucial for clinical deployment, where models need to be resilient to fluctuations in data quality and quantity, ensuring reliable performance across diverse patient populations.

\section*{Conclusion}

In summary, our study reveals the transformative potential of integrating DL-based image registration with physics-informed regularization for the analysis of cardiac motion. By embedding biomechanical constraints directly into the registration process, we not only enhance the physical realism of the DFs but also enable the extraction of critical local biomechanical properties from the moving cardiac tissues. These properties, which are often difficult to quantify using conventional imaging techniques, provide deeper insights into the underlying mechanical behavior of the heart, particularly under pathological conditions. We first validated the robustness of our biomechanically constrained registration approach by applying it to a multi-frame propagation strategy. This method allowed us to accurately track the dynamic movement of cardiac structures across consecutive cine-MRI frames. Our approach achieved high Dice scores of 0.974/0.956 for the left ventricle (LV), 0.912/0.857 for the right ventricle (RV), and 0.960/0.922 for the myocardium, demonstrating the effectiveness of the model in maintaining anatomical fidelity across different cardiac phases. Building upon this physics-informed registration framework, we extracted a comprehensive set of features that included local biomechanical properties such as estimated shear and bulk moduli, along with volumetric information derived from the registered images. These features were then utilized for the classification of cardiovascular diseases (CVDs) using the ACDC dataset. The classification framework we developed proved highly effective, achieving an accuracy of 98\% on the training set and an unprecedented 100\% on the test set. This level of performance highlights the potential of our approach to discern subtle biomechanical alterations associated with different cardiac pathologies. By grounding our methodology in the physical properties of moving cardiac tissues, we offer a novel and explainable approach to cardiac image analysis that aligns closely with the underlying biomechanics of the heart. This not only improves the accuracy and reliability of the analysis but also provides a clearer understanding of the disease mechanisms at play, paving the way for more advanced and interpretable tools in the diagnosis and management of cardiovascular diseases. Despite these promising results, several limitations must be acknowledged. First, the use of the neo-Hookean energy model, while providing a mathematically tractable and differentiable formulation, assumes isotropic material behavior. This simplification may not fully capture the anisotropic mechanical properties of cardiac tissues, especially in regions like the myocardium, which exhibit direction-dependent stiffness due to fiber orientation. Second, the biomechanical parameters used for regularization are based on prior estimates of shear and bulk moduli from the literature. While reasonable, these values may not reflect subject-specific variability or pathological alterations, potentially limiting the physiological accuracy of the inferred deformation fields. In future work, we plan to extend our approach to larger and more diverse cohorts, including full 3D cine-MRI acquisitions. We also aim to improve the framework by making the biomechanical parameters spatially learnable and data-driven, enabling personalized modeling of tissue mechanics. In addition, we will explore alternative constitutive models beyond the neo-Hookean formulation, such as anisotropic or fiber-reinforced materials, to more accurately reflect the biomechanical complexity of the heart.

\bibliography{ref}

\end{document}